\documentclass[final]{IEEEoj}
\usepackage{cite}
\usepackage{soul}
\usepackage{amsmath,amssymb,amsfonts}
\usepackage{subcaption}
\usepackage{array}
\usepackage{graphicx,color}
\usepackage{textcomp}
\def\BibTeX{{\rm B\kern-.05em{\sc i\kern-.025em b}\kern-.08em
		T\kern-.1667em\lower.7ex\hbox{E}\kern-.125emX}}
\AtBeginDocument{\definecolor{ojcolor}{cmyk}{0.93,0.59,0.15,0.02}}
\def\OJlogo{\vspace{-4pt}\hskip-4pt\includegraphics[height=18pt]{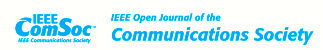}}

\begin{document}
	\receiveddate{XX Month, XXXX}
	\reviseddate{XX Month, XXXX}
	\accepteddate{XX Month, XXXX}
	\publisheddate{XX Month, XXXX}
	\currentdate{12 April, 2026}
	\doiinfo{OJCOMS.2026.XXXXXXX}
	
	\title{Secure Agentic AI and Multimodal Intelligence in 6G Communication Systems}
	
	\author{Satyam Gaba\IEEEauthorrefmark{1}\IEEEmembership{(Member, IEEE)},
		Devanshu Motwani\IEEEauthorrefmark{2},
		Siva Sai\IEEEauthorrefmark{3},
		Vineet Suri\IEEEauthorrefmark{4},
		Vikas Ranjan\IEEEauthorrefmark{4},
		Vinay Chamola\IEEEauthorrefmark{4}\IEEEmembership{(Senior Member, IEEE)}}
	
	\affil{Qualcomm Wireless Research, San Diego, CA 92121, USA}
	\affil{Department of Computer Science and Engineering, BITS Pilani, Pilani Campus, Rajasthan 333031, India}
	\affil{Department of Electrical and Computer Engineering, National University of Singapore 117583, Singapore}
	\affil{Department of Electrical and Electronics Engineering, BITS Pilani, Pilani Campus, Rajasthan 333031, India}
	\corresp{Corresponding author: Vinay Chamola (e-mail: vinay.chamola@pilani.bits-pilani.ac.in).}
	\markboth{Secure Agentic AI and Multimodal Intelligence in 6G Communication Systems}{Gaba \textit{et al.}}

	\begin{abstract}
		Sixth-generation (6G) communication systems are expected to support immersive, intent-driven, and latency-critical services through tight integration of multimodal intelligence, edge computing, and autonomous network control. In this setting, agentic AI systems can reason over heterogeneous inputs, use tools and network functions, coordinate across distributed entities, and adapt to dynamic environments. These capabilities make agentic multimodal intelligence promising for 6G applications such as semantic communications, digital twins, and zero-touch network management, but they also introduce security and privacy risks that extend beyond conventional AI and network security models. Vulnerabilities can propagate across multimodal data pipelines, memory, tools, communication interfaces, and inter-agent coordination loops. We survey multimodal intelligence and agentic AI for 6G cross-layered communication systems from the perspectives of privacy and security. We present a 6G-oriented system view spanning perception, memory, tool use, and multi-agent orchestration; analyze threat models and attack surfaces across sensing, cognition, action, memory, and communication layers; organize existing attacks into a structured taxonomy; and review mitigation strategies including prompt and memory defenses, inter-agent trust mechanisms, secure orchestration, and privacy-preserving learning. We conclude by outlining open research directions for trustworthy multimodal multi-agent 6G systems.
	\end{abstract}
	
	\begin{IEEEkeywords}
		6G, agentic AI, multimodal intelligence, multi-agent systems, network orchestration, wireless security, privacy, trustworthy AI, semantic communication
	\end{IEEEkeywords}
	
	\maketitle
	
	\section{Introduction}
	Sixth-generation (6G) communication systems are expected to move beyond high-throughput connectivity toward intelligent, immersive, and user-centric service delivery. Emerging applications such as holographic telepresence, tactile internet, extended reality, autonomous mobility, and digital twins require communication infrastructures that can jointly process heterogeneous data streams, adapt to rapidly changing environments, and support real-time decision-making under strict latency, reliability, and privacy constraints. Accordingly, 6G is increasingly envisioned as an intelligent communication fabric in which sensing, inference, orchestration, and actuation are tightly integrated~\cite{saad2020vision6g,letaief2019roadmap6g}.
	
	Meeting this vision requires both multimodal intelligence and agentic AI. Future 6G systems must fuse text, audio, video, sensor data, network telemetry, and contextual signals to infer user intent and optimize communication performance. At the same time, agentic components are increasingly deployed to support zero-touch management and autonomous network control~\cite{xie2021deepsc,yao2023react,park2023generativeagents}. However, coupling multimodal inputs, persistent memory, tool use, and inter-agent coordination broadens the attack surface beyond conventional AI pipelines or static network-control software.Vulnerabilities can propagate across multimodal channels, memory stores, orchestration interfaces, communication links, and control loops, leading to prompt-based manipulation, privilege abuse, data exfiltration, compromised orchestration decisions, and cascading failures that affect confidentiality, integrity, availability, latency, quality of experience (QoE), and even physical-world safety~\cite{greshake2023indirect,debenedetti2024agentdojo}.
	
	Although a growing body of work has examined LLM safety, agentic AI security, AI-enabled network automation, GenAI-enabled 6G, and future intelligent communications, current surveys still treat these topics through partially overlapping lenses~\cite{wang2024surveyagents,das2025llmsecuritysurvey,sheraz2025genai6g,durante2024agentai}. Existing studies typically emphasize one dimension of the problem, such as prompt injection in AI agents, adversarial attacks on multimodal models, GenAI applications in 6G, or security challenges in AI-native wireless networks, without centering the joint implications of multimodal inputs, persistent agent memory, tool authority, inter-agent trust, and network-control objectives. This paper takes a narrower and more specific position: it organizes architectural foundations, threat surfaces, attack taxonomy, and mitigation strategies around cross-layer security and privacy for multimodal, memory-enabled, tool-using, multi-agent 6G orchestration.
	
	Against this backdrop, the main contributions of this paper are as follows:
	\begin{itemize}
		\item We present a communication-centric survey framework that jointly treats single- and multi-agent LLM architectures, multimodal perception, persistent memory, tool use, and 6G communication-system integration as a single cross-layer security problem.
		\item We develop a formally specified threat model $\mathcal{S} = (\mathcal{A},\mathcal{B},\mathcal{E})$ over protected assets, trust boundaries, and adversary capability vectors, and tie each threat $\tau$ to an explicit communication-layer harm and to ITU-R IMT-2030 service metrics.
		\item We organize attacks into a four-class taxonomy --- prompt-hacking, socio-technical, adversarial/data-centric, and orchestration/communication-system --- grounded in demonstrated attack literature and mapped to 6G propagation scenarios for slice control, digital twins, and zero-touch management.
		\item We review cross-layer mitigation strategies, including prompt defenses, memory protection, secure tool use, inter-agent trust management, privacy-preserving learning, and resilient orchestration mechanisms.
		\item We outline open research challenges and future directions for trustworthy, privacy-preserving, and communication-efficient agentic AI systems in 6G networks.
	\end{itemize}
	
	The remainder of this paper is organized as follows. Section II provides background on agentic AI and multimodal intelligence in 6G systems. Section III positions this survey within the related literature. Section IV presents the threat model and attack surfaces. Section V discusses the attack taxonomy for secure agentic and multimodal 6G systems. Section VI reviews mitigation strategies and defense design principles. Section VII highlights open challenges and future research directions. Section VIII concludes the paper. Before turning to specific threats and defenses, the next section summarizes the 6G, multimodal AI, and agentic AI building blocks that make this problem space distinct.
	
	\begin{figure*}[t]
		\centering
		\includegraphics[width=\linewidth,trim=0cm 2cm 1cm 1cm,clip]{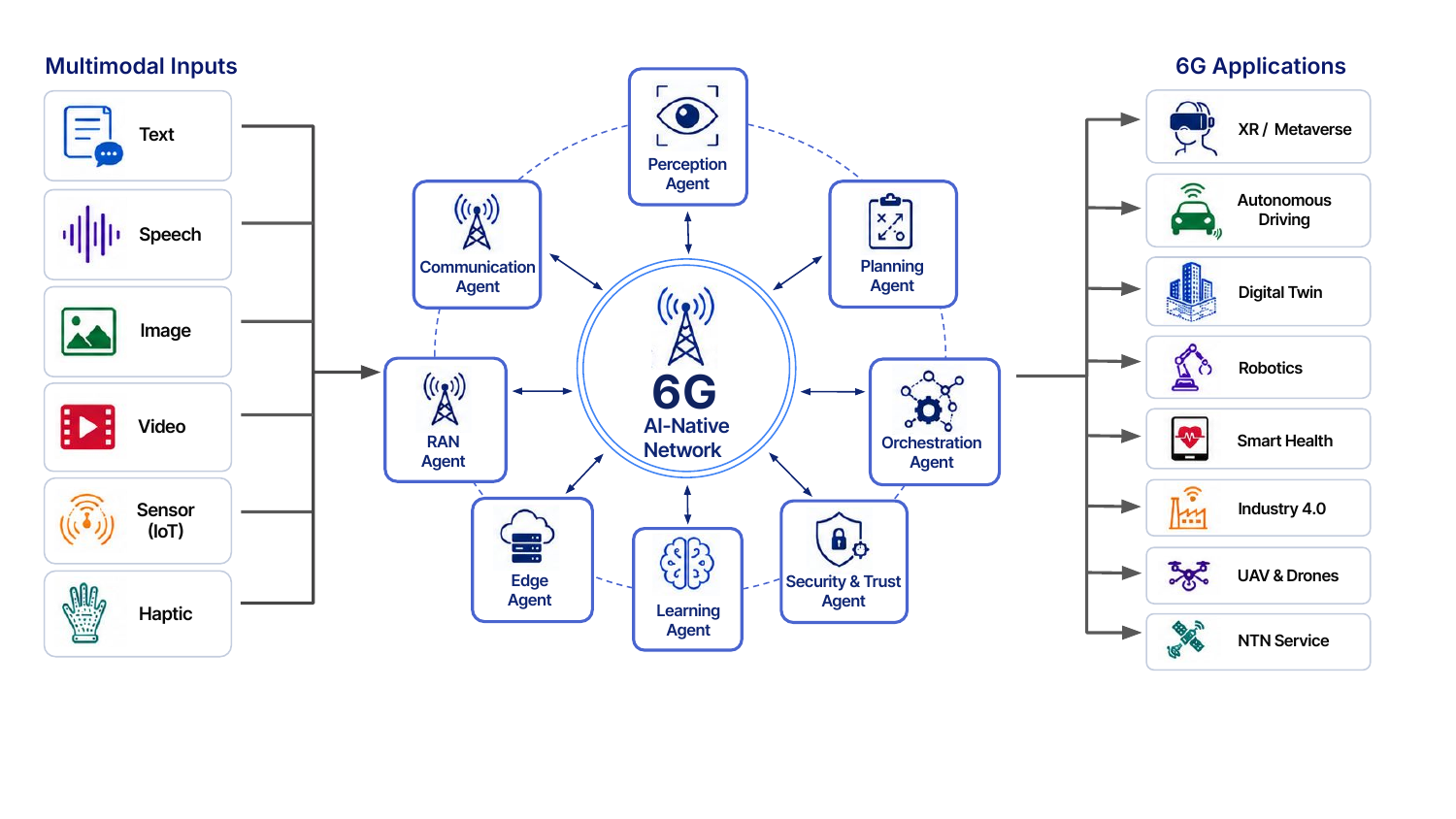}
		\caption{{High-level vision of multimodal, multi-agent agentic AI in 6G communication systems.}}
		\label{fig:6g_agentic_overview}
	\end{figure*}
	
	\section{Background on Agentic AI and Multimodal Intelligence in 6G Systems}

	\subsection{6G Communication Context}
	6G communication systems have prospects to unify communications, sensing, computing, and intelligence functions. These systems support services such as immersive extended reality, collaborative robotics, autonomous transportation, industrial automation, and digital-twin-assisted operations~\cite{saad2020vision6g,boutouchent2025intense,10092275}. 6G systems provide lower response times, higher reliability, and support for larger context windows compared to earlier-generation networks like 5G or 4G. They provide enhanced capability to adapt to user mobility, environmental dynamics, and joint communication-computation constraints~\cite{lu2021adaptiveedge}. This makes intelligent decision-making a basic requirement for robust communication systems.
	
	\subsection{Multimodal Intelligence in 6G}
	Multimodal systems functioning in the 6G realm allow a combination of text, audio, visual, haptic, and sensing inputs by extracting semantic features and aligning multimodal inputs~\cite{xie2021deepsc,chen2024crossmodal}. This becomes particularly relevant in systems focused on semantic-relevance-based communication, such as immersive communication or intelligent network management~\cite{cao2024mllmpsc,yang2025mgsc}.
	
	These multimodal capabilities underpin several concrete 6G services. In semantic and goal-oriented communication, the transmitter conveys task-relevant meaning rather than raw bits, and the receiver reconstructs the intended semantics for downstream decision-making, reducing traffic while preserving task utility~\cite{xie2021deepsc,gunduz2023semantic}. Integrated sensing and communication (ISAC) reuses shared spectrum, waveforms, and hardware for both data transfer and environment sensing, so that radar, lidar, telemetry, and visual streams are fused into a common operational state for closed-loop network control~\cite{liu2022isac,khan2026rfsensinggpt}. Immersive extended reality (XR) and tactile-internet services close real-time human-in-the-loop control loops over coupled visual and haptic feedback, imposing stringent latency and reliability requirements on the underlying multimodal pipeline~\cite{antonakoglou2018haptic,chen2024crossmodal}. However, this also expands the attack surface because corrupted or adversarial content in one modality may spread into system-wide decisions~\cite{shayegani2024jailbreakpieces}.

	\subsection{Agentic AI in Communication Systems}
	Throughout this paper, an \emph{agentic AI system} refers to an LLM-based system that couples a goal-directed reasoning loop with persistent memory and tool-invocation authority over external interfaces. An \emph{agent} denotes a single such system, or a specialized role within a multi-agent configuration. An \emph{agentic 6G system} denotes an agentic AI system embedded in a 6G control loop, where its actions feed back into communication-system state through orchestration, slice, or digital-twin interfaces. Agentic AI systems provide the autonomy to take actions based on the multimodal context. They enhance the capabilities of conventional AI components with planning, memory, tool use, and iterative execution~\cite{yao2023react,park2023generativeagents}. In communication systems, network knowledge and tools are used for multi-step problem-solving and semantically-guided orchestration~\cite{jiang2024commllm,boutouchent2025intense}. Agentic AI also become relevant in cloud-edge placement and digital-twin-assisted control in dynamic 6G environments~\cite{ayepahmensah2026ainative,li2025delayaware}.
	
	Fig.~\ref{fig:6g_agentic_overview}  frames 6G at high-level as a coordinated multi-agent system in which perception and planning agents process multimodal inputs, while communication, RAN, orchestration, edge, learning, and security-and-trust agents operate across network and service-control levels. This coordination enables services such as XR/metaverse experiences\cite{10296061}, autonomous driving, digital twins, robotics, smart health, Industry 4.0 automation, UAV/drone operations, and NTN services. From a systems perspective, a 6G agent is not a conversational agent alone, but a constrained observe-estimate-plan-act loop~\cite{boutouchent2025intense,ayepahmensah2026ainative}. As illustrated in Fig.~\ref{fig:6g_single_agent_architecture}, multimodal observations are first converted into an operational state comprising service intent, network telemetry, resource availability, and policy context. In the planning step, that state is mapped to candidate actions such as edge placement, semantic adaptation, or digital-twin-assisted control. In the next step, guardrails are enforced for access control, latency bounds, safety constraints, and protocol compliance, after which tool calls or digital twin queries trigger the execution of network functions. The resulting telemetry and action outcomes are then written back into memory for subsequent decisions. These memory operations occur across different timescales: fast local loops run near the device or edge, while slower supervisory loops handle cross-domain coordination and validation~\cite{shinn2023reflexion,li2025delayaware}.
	
	\subsection{Single-Agent Architecture}
	A truly autonomous system functions by decomposing the agent into its core functional modules. A single-agent architecture provides the basic building block for autonomous operation in 6G environments. Functionally, this combines perception, cognition, memory, action, and a tool interface based on the policy or guardrail constraints~\cite{yao2023react,schick2023toolformer}. In this setup, perception takes in multimodal context, such as user instructions, telemetry, and service state; cognition plans based on system objectives and constraints; memory preserves operational continuity; and action modules invoke network functions, such as querying digital twins, updating policies, or calling orchestration APIs. Because these modules connect heterogeneous inputs to system state and real-time actions, they form a relevant security boundary.
	
\begin{figure*}[t]
		\centering
		\includegraphics[trim=4cm 1cm 4cm 1cm, clip, width=\linewidth]{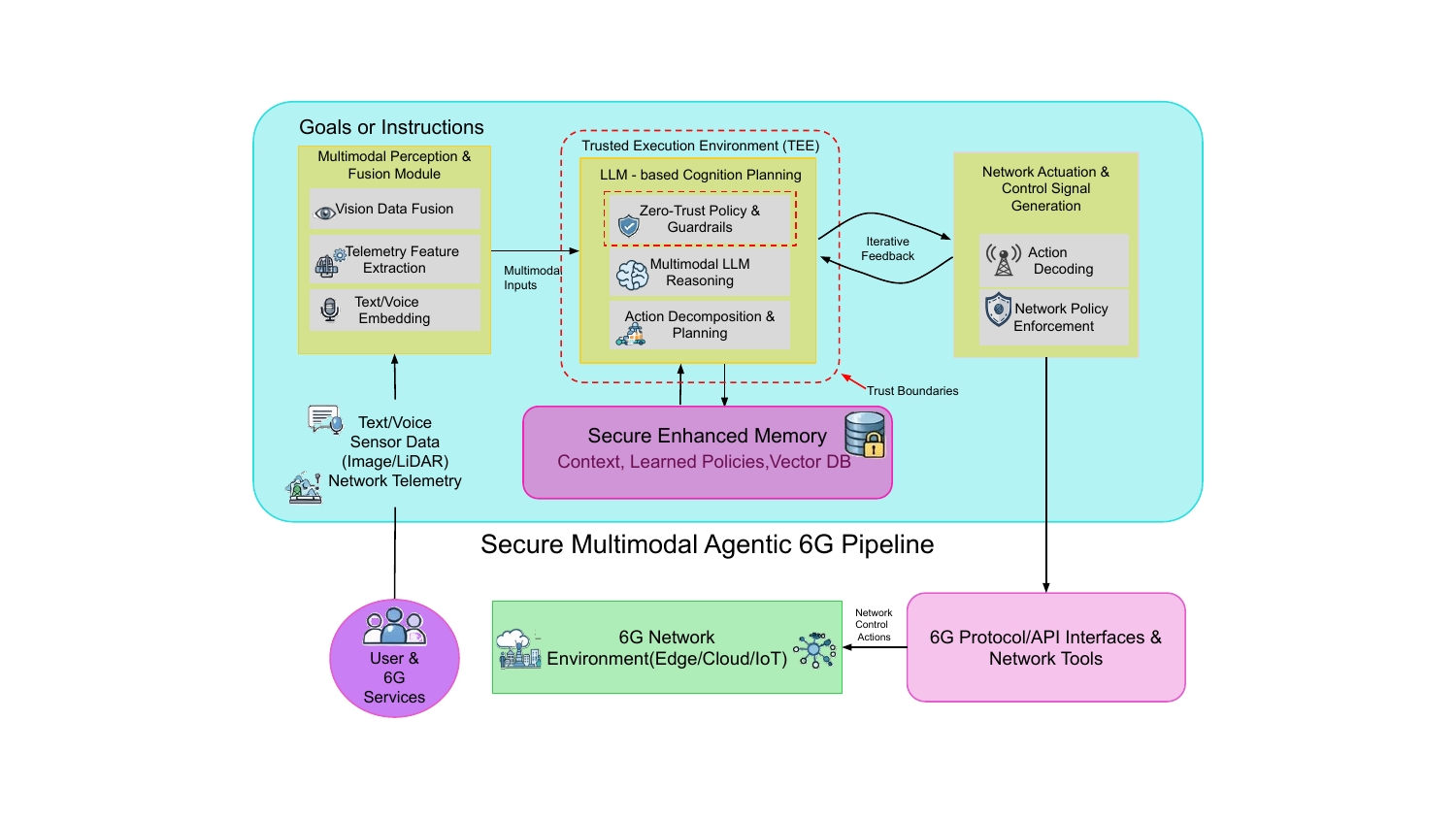}
		\caption{{ Single-agent orchestration for a 6G framework within a trusted execution environment.}}
		\label{fig:6g_single_agent_architecture}
	\end{figure*}

	As illustrated in Fig.~\ref{fig:6g_single_agent_architecture}, the single-agent pipeline integrates a trusted execution environment-bound LLM cognition module thus enforcing zero-trust policies across various multimodal inputs ranging from vision, telemetry, and text streams. The Secure Enhanced Memory preserves the context and learned policies across multiple reasoning cycles, while the action decoding module translates the planning outputs into verified network control signals. All interactions that occur with the 6G protocol or other API interfaces get routed through policy-checked, verifiable pathways, thus ensuring that all execution occurs within well-defined trust boundaries.
	
	\subsection{Multi-Agent Architecture} 
	A single agent, however, may not suffice for the safety requirements of a practical 6G network. Multi-agent architectures facilitate collaboration of specialized agents through coordination, delegation, and shared context across user devices, edge nodes, access networks, core infrastructure, service platforms, and digital twins~\cite{wu2024autogen,jiang2024commllm}. A multi-agent 6G system combines roles such as intent understanding, multimodal fusion, edge orchestration, slice management, digital twin support, and policy verification. The agent coordination can be managed centrally or through decentralized peer-to-peer exchange~\cite{ayepahmensah2026ainative}.
	
\begin{figure*}[t]
		\centering
		\includegraphics[width=\linewidth,trim=4cm 0 4.5cm 0,clip]{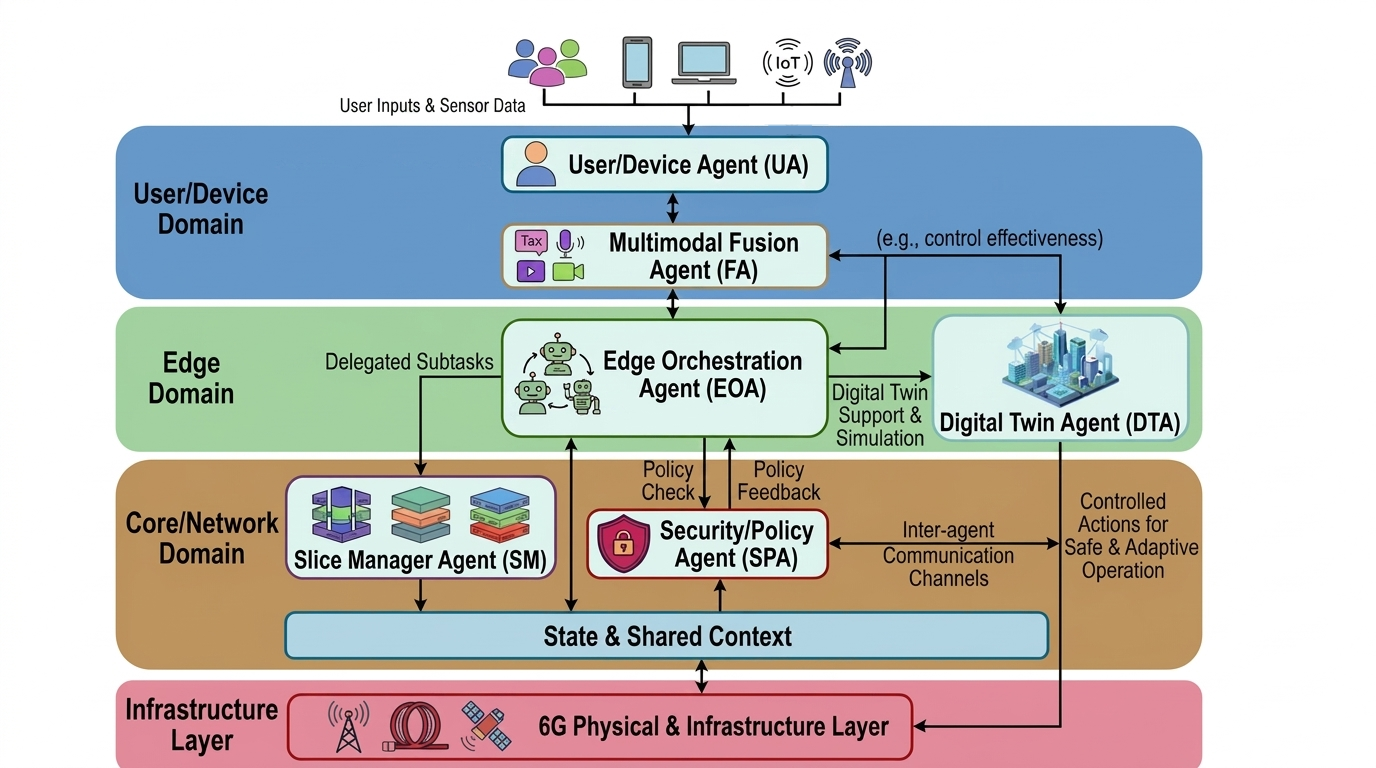}
		\caption{Multi-agent orchestration architecture for distributed 6G network control.}
		\label{fig:6g_multi_agent_architecture}
	\end{figure*}

	Fig.~\ref{fig:6g_multi_agent_architecture} depicts the coordination of assigned subtasks using structured communication channels across the Slice Manager, Security/Policy Agent, and Digital Twin Agent by the Edge Orchestration Agent. The Multimodal Fusion agent combines the user and sensor inputs into a unified context and passes it to the orchestration layer, whereas the Distributed Shared Memory maintains a consistent system state that is accessible to all agents in the system to perform reading or writing operations. This role differentiated design helps in enabling constrained adaptive control across the complete end-to-end 6G framework.
	
	Table~\ref{tab:llm_agentic_6g_comparison} summarizes the progression from conventional LLM systems to agentic AI systems and, finally, to multimodal multi-agent 6G systems. The comparison demonstrates how 6G systems go beyond LLM deployments: it couples persistent cross-domain state, network-facing tool authority, hierarchical coordination, and communication-performance constraints in ways that increase both system complexity and the security/privacy surface. This distinction provides the background for the upcoming related-work discussion and motivates the communication-centric threat framing used throughout the rest of the paper.
	
	\begin{table*}[t]
		\centering
		\caption{Comparison of LLM systems, agentic AI systems, and multimodal multi-agent 6G systems.}
		\label{tab:llm_agentic_6g_comparison}
		\renewcommand{\arraystretch}{1.15}
		\setlength{\tabcolsep}{4pt}
		\begin{tabular}{|p{2.4cm}|p{3.5cm}|p{3.8cm}|p{5.4cm}|}
			\hline
			\textbf{Dimension} & \textbf{LLM Systems} & \textbf{Agentic AI Systems} & \textbf{Multimodal Multi-Agent 6G Systems} \\
			\hline
			Primary function & Text generation and question answering & Goal-directed reasoning with tools and memory & Context-aware orchestration across communication, sensing, computing, and control layers \\
			\hline
			Input modalities & Mostly text & Primarily text with optional structured context & Text, audio, video, sensor streams, telemetry, location, and contextual metadata \\
			\hline
			State handling & Largely session-bounded & Persistent memory and task history & Persistent, distributed, and cross-domain state across device, edge, RAN, core, and digital twins \\
			\hline
			Tool interaction & Limited or optional & Direct API and tool invocation & Tool use tied to network functions, orchestration APIs, telemetry systems, and edge/cloud services \\
			\hline
			Coordination pattern & Single model response & Single-agent or simple chained workflows & Multi-agent, distributed, hierarchical, and latency-sensitive coordination \\
			\hline
			Failure propagation & Usually localized to a session & Can persist across tasks and tools & Can propagate across agents, network domains, slices, QoE pipelines, and control loops \\
			\hline
			Security focus & Prompt safety and model misuse & Memory integrity, tool safety, and policy alignment & Cross-layer security spanning multimodal inputs, memory, orchestration, trust, privacy, and communication performance \\
			\hline
			Recovery complexity & Re-prompt or reset & Rollback, memory repair, or policy update & Coordinated recovery across agents, edge/cloud domains, digital twins, and communication resources \\
			\hline
		\end{tabular}
	\end{table*}

	\section{Related Work and Survey Positioning}
	
	\subsection{Survey Methodology and Source Selection}
	This paper follows a systematized survey strategy rather than a formal PRISMA meta-review. The goal is to synthesize a rapidly evolving interdisciplinary literature into a communication-centric security framework, not to claim exhaustive bibliometric coverage. The corpus was compiled by searching IEEE Xplore, ACM Digital Library, arXiv, and Google Scholar with keywords from five seed areas: agentic AI security, multimodal safety, 6G and AI-native networking, federated and privacy-preserving learning, and multi-agent trust and orchestration. The temporal scope is 2019--2026, with earlier seminal works retained where they anchor a primitive still in use. The source corpus was organized around five seed areas. Foundational 6G and AI-native networking work was anchored in canonical vision and operations papers~\cite{liyanage2022zsm}. Agentic and multimodal foundations were anchored in primary framework papers and focused safety reviews~\cite{yao2023react,wu2024autogen,liu2024mllmsafety}. Attack and defense coverage was anchored in representative prompt-injection, evaluation, poisoning, and privacy works~\cite{greshake2023indirect,debenedetti2024agentdojo,biggio2012poisoning,dwork2006calibrating}. A work was included if it addressed at least one of: an attack, defense, benchmark, or deployment analysis for an LLM-based or agentic AI system; a 6G or AI-native networking capability relevant to the attack surface; multimodal perception, memory, or tool use in a cross-layer setting; or an adjacent subfield against which the paper is positioned in Table~\ref{tab:survey_positioning}. A work was excluded if it addressed purely hardware-layer or physical-layer security without an AI component, studied pre-LLM AI systems not applicable to prompt-driven agents, or was a weakly related preprint without primary evidence or archival follow-up. The final corpus comprises over~75 peer-reviewed works plus a small number of recent preprints covering emerging attacks.
	
	The inclusion strategy prioritized peer-reviewed journal and conference papers from the communications, networking, security, and trustworthy-AI communities that this survey bridges. Communications and networking venues such as IEEE Transactions on Wireless Communications, IEEE Transactions on Network and Service Management, IEEE Transactions on Communications, IEEE Communications Magazine, IEEE Network, IEEE/ACM Transactions on Networking, IEEE Communications Surveys \& Tutorials, IEEE Internet of Things Journal, IEEE INFOCOM, IEEE ICC, and IEEE GLOBECOM, supply the 6G, AI-native networking, semantic communication, zero-touch management, and digital-twin foundations on which the cross-layer threat model is built. Systems- and applied-security venues such as IEEE S\&P, ACM CCS, NDSS, USENIX Security, IEEE Transactions on Information Forensics and Security, and IEEE Transactions on Dependable and Secure Computing, supply the LLM-agent, prompt-injection, and adversarial-ML primary literature. General machine-learning venues such as NeurIPS, ICML, ICLR, IJCAI, and AISTATS, supply the multimodal-safety and agent-coordination foundations. Because the survey is positioned in a communications journal, communications and networking venues dominate the corpus, with security and ML venues included where they supply the attack or defense primitives that the 6G deployment scenarios build on. Restricting the corpus to any single community would under-represent at least one of the three cross-layer dimensions (communications, security, trustworthy AI). Recent preprints were included only when the topic is still emerging, and no mature archival literature is available, especially for newly observed multimodal and agentic AI attacks. The synthesis groups sources by system role rather than by publication venue: communication-system foundations, agentic and multi-agent architectures, attack surfaces, defense mechanisms, and open 6G deployment challenges. Where direct 6G evidence remains limited, the paper explicitly uses adjacent agent-security and multimodal-safety literature to motivate 6G threat-modeling extrapolations rather than presenting those propagation chains as already well established in deployed systems. This organization is intended to make the survey useful for both communications researchers and trustworthy AI researchers.

	\subsection{Surveys on Agentic AI and LLM Security}
	Recent surveys on LLM and agentic AI security provide the first set of papers against which this work is positioned. This literature covers broad security and privacy challenges of LLMs, including model misuse, data leakage, adversarial attacks, and system-level risks, and also surveys LLM-based autonomous agents by organizing the agent stack around profiling, memory, planning, and action modules~\cite{das2025llmsecuritysurvey,wang2024surveyagents}. More focused work examines the security and privacy of LLM agents directly, including threats, impacts, defenses, and case studies~\cite{he2025llmagentsecurity}.
	
	These papers are important because they establish the agentic AI security vocabulary used in this survey: prompt injection, jailbreaking, memory poisoning, tool misuse, data leakage, autonomous action risk, and unsafe inter-agent interaction. Their limitation for the present special issue is their deployment model. They primarily study general-purpose assistants, enterprise workflows, code/web agents, or cybersecurity operations, rather than agents embedded in latency-sensitive communication systems. They therefore do not systematically connect agent memory, tool authority, multimodal context, and multi-agent coordination to 6G-specific consequences such as slice misconfiguration, QoE degradation, edge-resource exhaustion, semantic distortion, radio-control instability, or digital-twin desynchronization.
	
	\subsection{Surveys on AI-Enabled 6G Systems}
	While the works above establish the agent-security baseline, a second body of literature moves into the communication-system setting that motivates this survey. Widely cited 6G vision papers identify AI, edge intelligence, ultra-low-latency services, and new application classes as central to 6G networks~\cite{saad2020vision6g,letaief2019roadmap6g,dang2020what6g}. Related surveys cover computational intelligence for 6G~\cite{ji2021computational6g}, federated learning and blockchain for digital-twin-empowered 6G edge networks~\cite{lu2021flblockchaindt,patel2025blockchain}, zero-touch network and service management for 5G and beyond~\cite{liyanage2022zsm}, digital twins for 6G~\cite{masaracchia2022digitaltwin6g}, and semantic communication networks with security and privacy considerations~\cite{guo2024semanticnetworks}. More recent work moves closer to the target of this survey by examining GenAI-enabled 6G technologies and applications and by providing a tutorial view of the transition from large AI models to agentic AI for future intelligent communications~\cite{sheraz2025genai6g,jiang2026agenticcommunications}.
	
	These surveys provide the communication-system foundation for this paper, but most treat AI as an optimization, prediction, management, generative modeling, or networking technology rather than as a security-critical, tool-using, autonomous agent with persistent memory and multi-step authority. Even works that explicitly discuss large AI models or agentic AI for future communications are primarily tutorials or technology surveys: they motivate agentic intelligence for networking, but do not provide a dedicated taxonomy of prompt-mediated control, agent-to-tool privilege boundaries, long-horizon memory poisoning, cross-agent trust exploitation, or multimodal prompt/context attacks in 6G control loops. As a result, the existing AI-for-6G literature does not fully capture how agentic autonomy changes the attack surface of zero-touch and semantic 6G systems.
	
	\subsection{Surveys on Multimodal Intelligence and Security}
	A third line of related work narrows further toward the multimodal setting, but still mostly treats safety at the model level rather than the system level. Existing surveys cover multimodal large language models, including architectures, training methods, datasets, and evaluation trends, and also review safety risks of multimodal large language models on images and text, including evaluation, attack, and defense methods~\cite{yin2024mllmsurvey,liu2024mllmsafety}. Related work also broadens the discussion to multimodal agent interaction in physical and virtual environments, emphasizing perception, embodiment, and context-aware agent behavior~\cite{durante2024agentai}. Semantic communication surveys connect multimodal communication architectures with security and privacy issues. In contrast, recent multimodal attack papers show that cross-modal adversarial content can steer multimodal language models at runtime~\cite{guo2024semanticnetworks,shayegani2024jailbreakpieces}.
	
	Most multimodal surveys and benchmarks focus on model behavior, visual-language alignment, image-text safety, or semantic communication architecture. They generally do not analyze how multimodal evidence is retained in persistent agent memory, how multimodal context affects tool invocation, how a compromised modality propagates through multi-agent delegation, or how cross-modal attacks translate into communication-layer harms such as latency violation, false QoE estimation, unsafe content adaptation, or incorrect orchestration decisions.
	
	\subsection{Research Gap and Positioning}
	Taken together, the related work above shows that the gap is not a lack of studies on LLM security, AI-enabled 6G, GenAI-enabled 6G, agentic intelligent communications, or multimodal safety individually, but that these streams stop at different system boundaries. LLM and agent-security surveys explain memory, tools, and prompt-driven attacks, but do not ground them in 6G service metrics and network-control loops~\cite{he2025llmagentsecurity,debenedetti2024agentdojo}. AI-for-6G and future-intelligent-communications surveys cover semantic communication, digital twins, zero-touch management, edge intelligence, and agentic AI opportunities, but usually do not treat multimodal prompt manipulation, persistent memory compromise, agent-to-tool authorization, and cross-agent trust as a unified 6G security problem~\cite{long2025llm6gintelligence,lu2021flblockchaindt}. Multimodal-intelligence surveys and attack papers explain multimodal fusion and model-level safety, but rarely connect multimodal compromise to inter-agent coordination, orchestration APIs, slice control, digital-twin state, or QoE-aware resource adaptation~\cite{liu2024mmsafetybench,bailey2024imagehijacks}.
	
	Classical machine-learning security studies for cyber-physical systems have also surveyed evasion, poisoning, privacy leakage, and defense mechanisms for sensor-, actuator-, and IoT-driven intelligent infrastructures~\cite{singh2022mlcpssecurity}. However, such works primarily address ML pipelines in conventional cyber-physical applications and do not model the agentic setting in which multimodal prompts, persistent memory, tool invocation, and inter-agent coordination can translate semantic compromise into network-control actions. Therefore, this survey treats ML-CPS security as an important baseline and extends it toward multimodal, memory-enabled, tool-using 6G orchestration.
	
	This paper is therefore positioned at the intersection of these streams as a focused survey of cross-layer security and privacy for multimodal, memory-enabled, tool-using, multi-agent 6G orchestration. Accordingly, the contribution is best understood as a narrative, communication-centric survey and threat-modeling synthesis for a fast-moving interdisciplinary topic. In this setting, a prompt or image-based manipulation is not merely a model-safety problem; it may alter semantic interpretation, contaminate memory, trigger a network-management tool, propagate to peer agents, and ultimately degrade QoE, reliability, latency, privacy, or digital-twin consistency. Table~\ref{tab:survey_positioning} summarizes this contrast across the most relevant prior survey directions.

	\begin{table*}[t]
		\centering
		\caption{Comparative positioning of this survey against representative prior works.}
		\label{tab:survey_positioning}
		\renewcommand{\arraystretch}{1.3}
		\small
		\begin{tabular}{|p{0.26\textwidth}|c|c|c|c|p{0.38\textwidth}|}
			\hline
			\textbf{Survey direction} & \textbf{Agentic} & \textbf{Multimodal} & \textbf{6G} & \textbf{Sec/Priv} & \textbf{Key limitation} \\
			\hline
			LLM security survey~\cite{das2025llmsecuritysurvey} 
			& Partial & Limited & No & Yes 
			& No agent memory, tools, multi-agent, or 6G focus. \\
			\hline
			
			LLM-agent security~\cite{wang2024surveyagents,he2025llmagentsecurity} 
			& Yes & Limited & No & Yes 
			& No multimodal 6G orchestration or communication metrics. \\
			\hline
			
			AI-enabled 6G surveys~\cite{saad2020vision6g,letaief2019roadmap6g,ji2021computational6g,liyanage2022zsm} 
			& Partial & Partial & Yes & Partial 
			& AI used as optimizer, not prompt-driven agents. \\
			\hline
			
			6G security \& blockchain/AI~\cite{wang2020securityprivacy6g,lu2021flblockchaindt,guo2024semanticnetworks,masaracchia2022digitaltwin6g} 
			& Limited & Partial & Yes & Yes 
			& No unified model of prompts, memory, trust, and tool abuse. \\
			\hline
			
			GenAI communications~\cite{sheraz2025genai6g,liu2025llmnetworking,jiang2026agenticcommunications} 
			& Yes & Partial & Yes & Partial 
			& No cross-layer security taxonomy for 6G agent loops. \\
			\hline
			
			Multimodal MLLM safety~\cite{yin2024mllmsurvey,liu2024mllmsafety} 
			& Partial & Yes & No & Partial 
			& Model-level only; no 6G orchestration analysis. \\
			\hline
			
			This survey 
			& Yes & Yes & Yes & Yes 
			& Cross-layer security framework for multimodal multi-agent 6G orchestration. \\
			\hline
		\end{tabular}
        \\[1mm]
        \makebox[\textwidth][l]{%
        \footnotesize
        \textit{Note:} ``Sec/Priv'' indicates coverage of security and privacy aspects in the survey.
        }
	\end{table*}
	
	\section{Threat Model and Attack Surfaces in Secure Agentic 6G Systems}
	\label{sec:threat_model}
	
	\subsection{System Assets}
	A threat model for secure agentic AI and multimodal intelligence in 6G systems begins by identifying the assets that couple agent cognition to communication-system control. In recent AI-native 6G architectures, those assets extend beyond links and credentials to include intent and telemetry inputs, semantic state, memory, tool bindings, orchestration policies, and digital-twin state used for closed-loop control~\cite{boutouchent2025intense,long2025llm6gintelligence}. Agent-oriented systems work likewise treat memory and tool interfaces as first-class operational components rather than incidental metadata~\cite{schick2023toolformer,shinn2023reflexion}.
	
	More concretely, the critical asset categories are as follows:
	
	\begin{itemize}
		\item \textbf{Multimodal perception inputs:} user prompts, sensor streams, video, audio, telemetry, and contextual metadata that feed semantic inference and network adaptation~\cite{chen2024crossmodal,cao2024mllmpsc}.
		\item \textbf{Memory and knowledge stores:} short-term context windows, persistent memory, vector databases, and RAG pipelines that preserve agent state across tasks~\cite{shinn2023reflexion}.
		\item \textbf{Tool interfaces and execution authority:} API bindings, orchestration hooks, and network-function interfaces through which agents translate decisions into system actions~\cite{jiang2024commllm}.
		\item \textbf{Agent identities and coordination state:} agent roles, trust metadata, inter-agent messaging channels, and shared policy state that govern delegation and collaboration~\cite{li2023camel}.
		\item \textbf{Network-control and digital-twin state:} scheduling parameters, slice configurations, QoE targets, and digital-twin synchronization state used for closed-loop optimization~\cite{paul2025quantumdrltwin,li2025delayaware}.
	\end{itemize}
	
	The compromise of any of these assets can affect not only confidentiality and integrity, but also delay, reliability, service continuity, and operational safety.
	
	\paragraph{Formal threat-model specification}
	To make the analysis in this section precise and to anchor the downstream attack taxonomy, we represent an agentic 6G system as the tuple $\mathcal{S} = (\mathcal{A},\mathcal{B},\mathcal{E})$, where $\mathcal{A} = \{a_1,\ldots,a_5\}$ is the ordered set of protected assets enumerated above, $\mathcal{B}$ is the set of trust boundaries across which adversarial state transitions can occur, and the adversary is described by $\mathcal{E} = (\kappa,\alpha,\mathcal{K})$ with capability level $\kappa \in \{\mathrm{passive},\mathrm{active},\mathrm{insider}\}$, access vector $\alpha \subseteq \{\mathrm{prompt},\mathrm{network},\mathrm{memory},\mathrm{tool},\mathrm{identity}\}$, and system knowledge $\mathcal{K} \subseteq \mathcal{S}$. A threat is a triple $\tau = (\mathcal{E}_i, a_j, \delta)$ mapping an adversary--asset pair to an impact $\delta$ that manifests as one of several communication-layer harms: confidentiality or integrity loss, QoE degradation, slice or policy misconfiguration, latency or reliability violation, digital-twin desynchronization, or service-continuity loss. At runtime, an agent realizes a policy-constrained action $a_t = \pi\!\left(o_t, m_t, \mathcal{T}; \theta\right)$ that maps multimodal observations $o_t$, memory state $m_t$, and available tools $\mathcal{T}$ to network-facing actions under guardrails $\theta$; adversarial transitions in $\mathcal{B}$ corrupt one or more arguments of $\pi$ and propagate along this dynamics. Table~\ref{tab:threat_model_summary} enumerates the principal threats $\tau$ in this model, Table~\ref{tab:6g_attack_scenarios} shows how selected threats propagate end-to-end through $\pi$ in representative 6G deployments, and the attack taxonomy in Section~\ref{sec:attack_taxonomy} organizes those threats by primary entry into $\pi$.
	
	\subsection{Trust Boundaries}
	Protecting these assets requires understanding where trust changes across the system. In agentic 6G deployments, trust no longer shifts only at a network perimeter; it also shifts whenever untrusted content is interpreted by an agent, written into memory, or converted into tool calls~\cite{rose2020zerotrust,greshake2023indirect}. Because these systems rely on multimodal inputs, memory access, tool use, and distributed agent interaction, they operate across multiple interfaces where assumptions about trust may be violated. Threat modeling must therefore distinguish not only between internal and external entities, but also between trusted, partially trusted, and dynamically trusted interactions.
	
	Several trust boundaries are especially important. The user-to-agent boundary is exposed by indirect prompt injection and malicious-content attacks~\cite{zhan2024injecagent}. The agent-to-memory boundary is exposed because memory is vulnerable to malicious access and modification, thereby impacting the agent's future decisions, even though the base model remains the same.~\cite{chen2024agentpoison,zou2024poisonedrag}. The agent-to-tool boundary is critical because prompt-manipulated agents can misuse APIs, delegated actions, and orchestration interfaces~\cite{debenedetti2024agentdojo}. The agent-to-agent boundary matters in distributed settings, where a compromised peer can propagate false context or harmful behavior across a larger workflow~\cite{he2025communicationattacks,shahroz2025agentsundersiege}. The edge-cloud and network-domain boundary remains important because policy synchronization and digital-twin coordination span devices, edge nodes, and orchestration planes~\cite{lu2021flblockchaindt,paul2025quantumdrltwin,10296061}. Finally, the physical-to-digital boundary is relevant whenever sensed signals are fused into control-relevant semantic context, especially in multimodal or digital-twin pipelines~\cite{cao2024mllmpsc,khan2026rfsensinggpt}. Each of these boundaries can be exploited to inject false context, escalate privileges, corrupt decisions, or trigger cascading failures.
	
	\subsection{Adversary Model}
	Those boundaries define the adversary capabilities that the threat model must account for. The adversary model in secure agentic 6G systems must therefore account for both conventional cyber attackers and attack methods already demonstrated against LLM agents, such as indirect prompt injection, tool-triggered misuse, and memory poisoning~\cite{he2025llmagentsecurity}. Multimodal jailbreaks and cross-agent propagation extend the same threat picture to richer perception and distributed-control settings~\cite{bailey2024imagehijacks,gu2024agentsmith}. Adversaries may be external, internal, or partially trusted. Their goals may include stealing sensitive information, degrading service quality, manipulating orchestration outcomes, exhausting resources, inducing unsafe behavior, or covertly altering long-horizon system decisions.
	
	Representative adversaries are end users who alter outputs intended for genuine agents using malicious techniques. These include prompt injections, deceptive multimodal inputs, compromised tools, or malicious external data sources. This also includes insider attacks using privileged accounts to access memory stores or orchestration interfaces~\cite{zhan2024injecagent,shayegani2024jailbreakpieces}. A compromised agent acts as an adversary by manipulating delegated decisions while participating in multi-agent workflows~\cite{triedman2025mascode}. In addition, weaknesses in tool configuration or trust assumptions are exploited by adversaries. Persistent memory and long context window-based planning also expose opportunities for adversaries to cause harm to the system~\cite{wang2024raccoon,chen2024agentpoison}.
	
	Adversaries act in different ways, including generating user prompts, altering retrieved data, modifying external data stores, or maliciously using tools~\cite{greshake2023indirect,chen2024agentpoison}. In distributed deployments, the same adversary may also spoof agent identities or intercept inter-agent communication~\cite{triedman2025mascode}. In a stronger threat setting, the adversary may manipulate multimodal inputs such as visual frames, audio commands, or sensing streams~\cite{bailey2024imagehijacks,shayegani2024jailbreakpieces}. These capabilities distinguish agentic 6G threat models from conventional network attacks because the attacker can influence not only packet-level behavior, but also the semantic and cognitive processes that guide autonomous decisions~\cite{zhang2017dolphinattack}.
	
	\begin{figure*}[t]
		\centering
		\includegraphics[trim=1cm 0cm 3cm 0cm, clip, width=\linewidth]{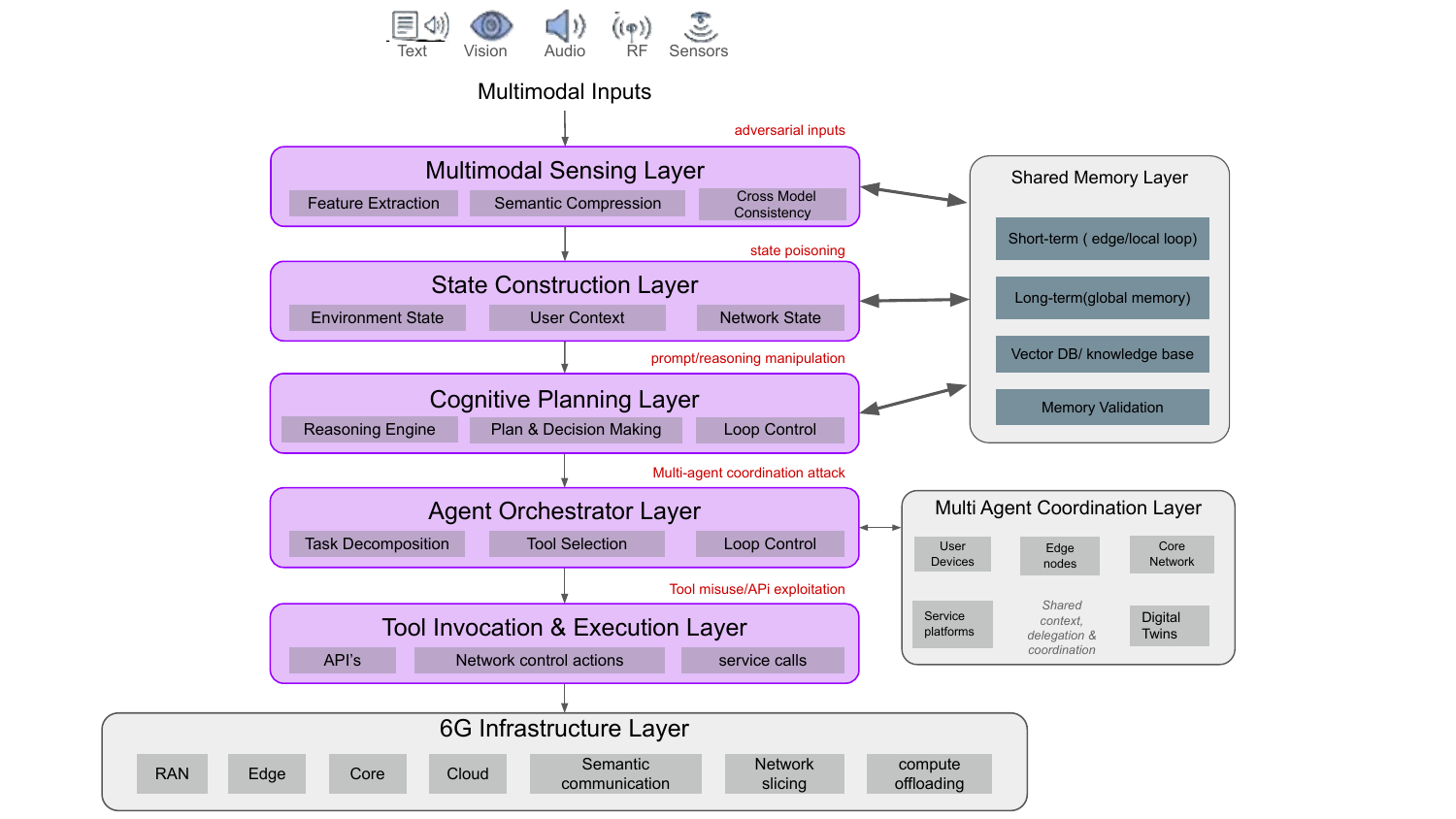}
		\caption{Layered architecture and cross-layer attack surfaces in a multimodal agentic 6G system.}
		\label{fig:6g_attack_surfaces}
	\end{figure*}
	
	\subsection{Attack Surfaces}
	Given those capabilities, the next step is to map where attacks can actually enter the system. The attack surface of a secure agentic 6G system can be organized into several interacting layers. A layered view is useful because attacks in one layer often propagate to others through memory updates, tool invocations, and control decisions. This section groups attack surfaces into multimodal input, reasoning, memory, tool and action, communication, and inter-agent coordination layers. The first attack surface is the multimodal sensing and input layer, which includes user prompts, retrieved documents, visual streams, audio signals, telemetry, and contextual metadata. Because these inputs are often heterogeneous and externally sourced, they create opportunities for indirect prompt injection, adversarial visual inputs, cross-modal manipulation, and semantic misdirection that have been demonstrated in both tool-integrated agents and multimodal LLMs~\cite{carlini2016hiddenvoice,shayegani2024jailbreakpieces}. In a 6G system, such attacks may alter the agent's understanding of user intent, network conditions, or service requirements, leading to downstream errors in adaptation or control.
	
	Another attack surface is the agent’s reasoning layer - analyze inputs, create plans, and select suitable tools. Compromises at this layer can occur through jailbreaks, changes in goals, immature planning, and improper resolution of conflicting contexts~\cite{zou2023universal,debenedetti2024agentdojo}. Such failures affect resource allocation, policy updates, service prioritization, and anomaly response, significantly impacting the system. A small change in the reasoning layer can have a major impact on the system's functionality. Information and memory form another vulnerable attack surface. The persistence of user preferences in memory helps with personalization and long-term optimization, but it can also be exploited. This also includes context window data, persistent memory stores, vector databases, and RAG pipelines~\cite{park2023generativeagents}. Red teaming reports suggest that manipulation of memory states can bias an agent’s behavior in malicious ways. This is true both for long-term memory as well as retrieval-based knowledge sources~\cite{chen2024agentpoison}. In the context of 6G communication systems, this results in inconsistent resource allocation, corruption of user profiles, modification of context information, and unauthorized policy modification~\cite{zou2024poisonedrag,xue2024badrag}.
	
	Agent tools are the next sensitive area for external attacks. Agents use tools to interact with data stores, network management APIs, digital twins, edge devices, logging, and monitoring frameworks. Tool usage is a critical step since it connects the reasoning layer to actual system actions. Unauthorized tool use, privilege escalation, malicious tool responses, or unsafe execution chains are different ways in which the attacker can change the policies, disrupt services, alter network behavior, or steal data~\cite{zhan2024injecagent,debenedetti2024agentdojo}. The autonomous nature of tool selection makes this step more vulnerable to agentic AI systems than to traditional software pipelines~\cite{ruan2023toolemu,hu2026maltool}. Inter-agent communication in a multi-agent setup forms the next attack surface under consideration. In multi-agent architectures, agents exchange messages, share context, delegate subtasks, and coordinate on shared memory or policies. This creates opportunities for identity spoofing, trust exploitation, message tampering, inconsistent state propagation, and cascading hallucinations or misaligned decisions~\cite{he2025communicationattacks,triedman2025mascode}. In 6G settings, where intelligence may be distributed across devices, edge nodes, and core-network functions, such attacks can spread across operational domains, leading to large-scale coordination failures.
	
	Taken together, these layers should be viewed as a cross-layer dependency graph rather than as isolated interfaces. An attack entering through multimodal input, reasoning, memory, tools, or inter-agent communication can propagate across the autonomous communication stack through state updates, delegated actions, and network-facing control loops. Fig.~\ref{fig:6g_attack_surfaces} summarizes these cross-layer attack entry points and the main paths by which local compromises can become system-level 6G failures.
	
	\begin{table*}[ht]
		\centering
		\caption{6G attack propagation scenarios for secure multimodal multi-agent systems.}
		\label{tab:6g_attack_scenarios}
		\renewcommand{\arraystretch}{1.4}
		\setlength{\tabcolsep}{4pt}
		\scriptsize
		\begin{tabular}{|
				>{\raggedright\arraybackslash}p{2.3cm}|
				>{\raggedright\arraybackslash}p{2.5cm}|
				>{\raggedright\arraybackslash}p{3.0cm}|
				>{\raggedright\arraybackslash}p{2.7cm}|
				>{\raggedright\arraybackslash}p{2.7cm}|
				>{\raggedright\arraybackslash}p{2.6cm}|
			}
			\hline
			\textbf{Scenario} & \textbf{Entry point} & \textbf{Propagation path} & \textbf{Consequence in Network} & \textbf{Containment direction} & \textbf{Violated 6G Metric} \\
			\hline
			Zero-touch slice-management prompt injection~\cite{greshake2023indirect} & Malicious inputs in tickets, logs, or user requests & Agent misinterprets the intent, leading to wrong slice/resource decisions & Wrong slice allocation, QoS drop, policy violation & Input sanitization, intent validation, least-privilege, human override & Slice SLA integrity; IMT-2030 sub-1\,ms URLLC latency~\cite{iturIMT2030} \\
			\hline
			Cross-modal QoE spoofing (immersive apps)~\cite{carlini2018audio} & Adversarial visual, audio, gaze, or context signals & Faulty fusion leads to semantic misinterpretation & QoE degradation, latency mismatch, poor adaptation & Cross-modal validation, anomaly detection, fallback policies & Immersive-service QoE; IMT-2030 uplink/downlink experience targets~\cite{iturIMT2030} \\
			\hline
			Poisoned digital-twin state (edge control)~\cite{tolpegin2020datapoison} & Tampered telemetry or historical data & Corrupted state leading to faulty predictions & Resource misallocation, instability, repeated control errors & Authenticated data, state consistency checks, rollback & Closed-loop control reliability; 3GPP Rel-18 ZSM consistency~\cite{liyanage2022zsm} \\
			\hline
			Inter-agent trust compromise~\cite{gleave2020adversarial,standen2025adversarial} & Spoofed agent identity or messages & False context leading to cascading coordination errors & Conflicting decisions, service disruption, unsafe actions & Agent authentication, message integrity, trust scoring, isolation & Orchestration availability; 99.999\% reliability for critical services~\cite{iturIMT2030} \\
			\hline
		\end{tabular}
	\end{table*}

	\subsection{Threat Categories}
	%   memory poisoning, cascading hallucinations, tool abuse, privilege escalation,
	%   data exfiltration, identity spoofing, coordination failures, DoS
	To simplify analysis, the layered attack surfaces above can be grouped into four high-level categories: memory and reasoning integrity threats, tool and privilege abuse threats, agent trust and deception threats, and coordination and availability threats. This abstraction is useful because different adversarial attacks often have similar impacts on the system, even though they may target different layers. 6G threat models should be capable of analyzing harms at the level of the communication system, like QoE degradation, latency, or reliability violations, policy inconsistency, and unsafe control behavior, besides the core aspects like confidentiality, integrity, and lack of availability. In practice, such attacks often surface first as 6G operational symptom such as orchestration instability, unsafe configuration changes, or service-availability drops. These symptoms attributed to an underlying security compromise once correlated with telemetry and audit logs, which has implications for how monitoring and incident response are structured. Table~\ref{tab:threat_model_summary} summarizes the main threat classes by linking each one to its targeted asset, the 6G layer it affects, and the system-level consequence it can produce, such as decision drift, trust collapse, privacy leakage, or denial of service.

	\subsection{Specific 6G Attack Propagation Scenarios}
	Table~\ref{tab:survey_positioning} presents a comparative analysis of this survey against representative prior works across key dimensions, which include agentic capability, 6G focus, and security/privacy coverage. It highlights that, unlike existing research studies, our work uniquely provides a unified, cross-layer security framework for multimodal multi-agent 6G orchestration with an extensive consideration of tools, memory, and communication aspects.
	Table~\ref{tab:6g_attack_scenarios} presents representative attack-propagation scenarios for secure multimodal multi-agent 6G systems, showing how each scenario has been demonstrated as an attack primitive in an adjacent domain and how it is expected to propagate through agentic workflows to produce 6G network-level consequences, motivating corresponding containment strategies. Prompt injection and tool misuse are directly supported by recent agent-security literature~\cite{greshake2023indirect,shi2025toolhijacker}. Memory poisoning and multimodal adversarial manipulation are likewise established in recent technical work~\cite{chen2024agentpoison,zou2024poisonedrag}. Adversarial audio and cross-modal attacks on perception systems~\cite{carlini2018audio}, federated-learning data poisoning and backdoor attacks against distributed training pipelines~\cite{tolpegin2020datapoison,bagdasaryan2020backdoorfl}, and adversarial-policy and inter-agent manipulation in multi-agent reinforcement learning~\cite{gleave2020adversarial,standen2025adversarial} have each been demonstrated experimentally in adjacent domains that share the relevant components of the 6G agentic stack. Their end-to-end impact on slice control, QoE, and digital twins therefore constitutes a grounded extrapolation that motivates the threat model developed here and that warrants dedicated empirical investigation in 6G testbeds~\cite{xu2024llmagents6g,lu2021flblockchaindt}.

	\begin{table*}[t]
		\centering
		\caption{Threat model summary for agentic and multimodal 6G systems.}
		\label{tab:threat_model_summary}
		\renewcommand{\arraystretch}{1.2}
		\setlength{\tabcolsep}{4pt}
		\begin{tabular}{|
				>{\raggedright\arraybackslash}p{2.4cm}|
				>{\raggedright\arraybackslash}p{4.2cm}|
				>{\raggedright\arraybackslash}p{2.7cm}|
				>{\raggedright\arraybackslash}p{2.6cm}|
				>{\raggedright\arraybackslash}p{3.8cm}|
			}
			\hline
			\textbf{Threat} & \textbf{Description} & \textbf{Targeted Asset} & \textbf{6G Layer Impacted} & \textbf{System Consequence} \\
			\hline
			Memory poisoning & Malicious context is inserted into persistent or short-term memory to bias future decisions & Memory stores, retrieval pipelines, digital twin state & Edge/cloud intelligence, orchestration, service state & Persistent misconfiguration, incorrect personalization, and long-horizon decision drift \\
			\hline
			Tool misuse & agent is induced to invoke unsafe or unintended APIs, tools, or workflows & Tool interfaces, network APIs, execution permissions & Management plane, orchestration interfaces, service platforms & Unauthorized actions, workflow disruption, and operational instability \\
			\hline
			Privilege escalation & Agent exceeds intended authority through manipulated prompts, policies, or chained actions & Guardrails, authorization context, policy state & Control plane, slice management, policy enforcement & access to restricted resources, unsafe reconfiguration, and policy violations \\
			\hline
			Cascading hallucinations & Incorrect outputs are reused by downstream agents or tools as trusted facts & Inter-agent context, reasoning state, shared knowledge & Multi-agent coordination, decision loops & Propagated errors, conflicting actions, and degraded QoE \\
			\hline
			Data exfiltration & Sensitive information is leaked through apparently legitimate communication channels & User data, telemetry, memory, policy data & Communication services, data plane, monitoring systems & Privacy loss, intellectual property leakage, and regulatory exposure \\
			\hline
			Coordination exploitation & A malicious or compromised agent manipulates distributed collaboration and task handoffs & Agent identity, messaging channels, shared state & Device-edge-cloud coordination, orchestration fabric & Cross-domain failures, trust collapse, and inconsistent system behavior \\
			\hline
			Resource exhaustion & Attackers trigger excessive reasoning, delegation, or tool usage to consume resources & Compute budget, API quotas, orchestration pipelines & Edge runtime, service orchestration, network control loops & Increased latency, service degradation, and denial of service \\
			\hline
		\end{tabular}
	\end{table*}
	
	\section{Attack Taxonomy for Secure Agentic and Multimodal 6G Systems}
	\label{sec:attack_taxonomy}
	Building on the formal model $\mathcal{S} = (\mathcal{A},\mathcal{B},\mathcal{E})$ from Section~\ref{sec:threat_model}, this section organizes how the threat instances $\tau$ manifest as concrete attacks on the agent policy $\pi$ across four classes. The threat model above identifies where harm enters; this section organizes how those threats manifest in practice. Unless otherwise noted, the attack primitives discussed below are established in adjacent AI-security and multimodal-safety literature, whereas their end-to-end consequences for 6G orchestration should often be read as communication-system extrapolations that motivate threat modeling rather than as already standardized deployment evidence.
	
	The four categories are organized by primary attack vector: prompt-hacking, socio-technical, adversarial and data-centric, and orchestration and communication-system. Some attacks span multiple categories; for example, multimodal jailbreaks involve both prompt-level instruction bypass and adversarial input perturbations. Such attacks are placed in the category that best describes their primary entry mechanism and cross-referenced where relevant. This placement is intended to aid navigation rather than to imply mutual exclusivity between categories.
	
	\subsection{Prompt-Hacking Attacks}
	Prompt-hacking attacks target the semantic interface through which agentic systems interpret goals, instructions, and contextual information~\cite{greshake2023indirect,debenedetti2024agentdojo}. In secure agentic 6G systems, these attacks are especially dangerous because natural language directives, multimodal prompts, policy descriptions, retrieved documents, and operator instructions may all influence downstream reasoning and control. Unlike conventional network attacks that exploit protocol weaknesses directly, prompt-hacking attacks manipulate the agent's internal interpretation of system objectives, causing the system to behave incorrectly while appearing syntactically valid~\cite{zhan2024injecagent,jiang2024commllm}. A primary class within this category is prompt injection, in which an attacker embeds malicious instructions into user inputs, retrieved content, documents, telemetry descriptions, or multimodal context. This attack aims to change the agent’s behavior by making unsafe API calls, changing orchestration priorities, distorting the interpretation of network conditions, or inducing the agent to reveal sensitive policy or system state. When the agent derives information from external sources or when the prompt context influences multiple autonomous modules, the threat from such prompt injections becomes all the more pronounced~\cite{zhan2024injecagent,wang2024raccoon}.
	
	Jailbreaking refers to attempts to bypass the behavioral guidelines, guardrail policies, and alignment restrictions designed for the agent. In the case of agentic architecture, jailbreak can compromise access to high-value tools, issue unauthorized actions, or weaken safety checks across an entire execution loop. In a 6G multimodal setup, this can exploit combinations of text, visual, and contextual inputs to avoid detection and lead to decisions that violate communication policies or operator constraints~\cite{zou2023universal,shayegani2024jailbreakpieces}. Tool chain manipulation is the mechanism by which an attacker attacks the agent's ability to invoke and sequence external tools. In such cases, adversaries manipulate the output of one tool, which then becomes the input to another tool. So, semantically, chaining seems correct; operationally, the behavior is significantly altered. In 6G communication systems, such attacks lead to unsafe orchestration pipelines, where a compromised retrieval step can result in faulty reasoning, triggering reconfiguration actions or data leakage through agent outputs. Such tool chaining contamination can corrupt multiple agents in a multi-agent scenario and gain illegitimate control over systems~\cite{debenedetti2024agentdojo,shi2025toolhijacker}.
	
	Taken together, prompt-hacking attacks exploit the fact that autonomous communication systems increasingly depend on semantic context rather than just structured control inputs. Their danger lies in the ability to convert high-level language or multimodal content into real system actions, making them a central risk category for secure agentic 6G deployment.
	
	\subsection{Socio-Technical Attacks}
	Beyond direct prompt manipulation, similar attacks can also exploit the human and organizational context around the agent~\cite{greshake2023indirect,park2023generativeagents}. These attacks are important in 6G systems because intelligent communication infrastructures will often operate in partially autonomous and human-supervised modes, especially in orchestration, fault management, policy approval, and service, thereby affecting the result. Attackers may target not only computational components, but also the trust relationships and behavioral assumptions that shape human-agent collaboration~\cite{chen2024agentpoison,ayepahmensah2026ainative}.
	
	One important class is social engineering against agent-mediated workflows. Here, the attacker uses deceptive content, persuasive instructions, or fabricated context to influence a human operator through the agent or to influence the agent through a human intermediary. In zero-touch and human-in-the-loop 6G systems, this may take the form of deceptive alarms, forged operational narratives, misleading summaries of network state, or crafted escalation requests that pressure operators into approving unsafe actions. Because agentic systems can generate highly customized and context-aware interactions, such attacks may become more scalable and harder to detect than traditional phishing or deception campaigns~\cite{zhan2024injecagent,wang2024raccoon}.
	
	Memory tempering is the second class of socio-technical attacks. An agent stores information for taking personalized future decisions, and a manipulation at this level introduces bias in the agent’s output. This is achieved by inserting manufactured preferences, changing incident histories, or modifying retrieval results so that the agent makes harmful decisions over time. In 6G systems, memory tampering embeds false assumptions and thereby affects personalized responses, QoE adaptation, anomaly response, or resource scheduling. The danger is that it affects the agent's autonomous behavior in a sustained way through context manipulation. And it is difficult to identify since it appears within a legitimate learned context~\cite{chen2024agentpoison,zou2024poisonedrag}.
	
	Trust exploitation is another area where the system behavior can be influenced significantly. This breach of trust can occur at the level of a human or a peer agent. In 6G networks, humans depend on the recommendations and summaries generated by a compromised agent, or a specialized agent may trust the inputs of a peer agent without separately validating their authenticity. This attack is more likely to affect systems that are not fully evolved in terms of explainability or automation.~\cite{triedman2025mascode,shahroz2025agentsundersiege}
	
	\subsection{Adversarial and Data-Centric Attacks}
	Data-centric attacks are the ones that target the data determining the learned behavior of the agent~\cite{biggio2012poisoning,tramer2016stealing}. It includes data sources such as telemetry, sensor stream inputs, operator selections, and dynamic service data. These attacks exploit the dependence of multimodal agentic systems on external data sources, retrieved knowledge, contextual memory, and learned model behavior. Such attacks can degrade service quality or even lead to persistent system-level failures. Data poisoning~\cite{biggio2012poisoning} is another such attack, in which the training data, retrieval information, and data stores are manipulated by injecting fabricated data. In this way, the attacker biases the model’s behavior without actually touching the reasoning layer. In a 6G scenario, data poisoning occurs in network telemetry repositories, incident databases, digital twin histories, or multimodal knowledge used for orchestration and diagnosis. Such poisoning of data can misdirect agents to make decisions that lead to poor resource-allocation decisions, misclassifying faults, or prioritizing incorrect services~\cite{zou2024poisonedrag,xue2024badrag}.
	
	Backdoor attacks constitute another class, in which the malicious behavior is event-triggered. This means that the agent behaves normally on regular inputs, but when specific inputs are triggered, it generates outputs under the adversary's control. In practical 6G systems, this backdoor is triggered by an unusual telemetry reading, a specific operator command, or a particular document structure. And this can manipulate an agent into executing unsafe actions, revealing sensitive information, or disabling guardrails. The hidden nature of backdoor attacks makes it difficult to predict them in an autonomous, agentic system~\cite{gu2019badnets,liu2018trojaning,huang2024compositebackdoor}. Model extraction and functionality theft work in different ways. The attacker repeatedly queries the system to understand response patterns and gradually decipher the model's functionality based on them. Over time, they can approximate the decision logic and create a functional clone of the system. In 6G systems, in addition to the risk of exposing the model, adversaries can also access orchestration patterns, memory-dependent behaviors, and policy selection strategies. This information can be used for more targeted attacks, including evasion, inversion, and policy abuse. Commercially deployed edge systems are especially exposed because their external interfaces are, by necessity, accessible~\cite{tramer2016stealing,fredrikson2015modelinversion}.
	
	In multimodal adversarial manipulation, the attackers don’t steal information or inject triggers, but actually perturb the multimodal inputs in such a way that the disturbance is not visible to the human eye, and yet it is enough to confuse the agent’s reasoning. In 6G multimodal deployments, attackers can affect any modality, such as visual streams, spoken commands, sensor readings, or the fused context representations that drive semantic communication. The assumption of cross-modal consistency becomes the weakness of the reasoning engine, since manipulation of any of the modalities can induce inconsistent or unexpected outputs~\cite{carlini2017robustness,shayegani2024jailbreakpieces, liu2024mmsafetybench}.
	
	\subsection{Orchestration and Communication-System Attacks}
	Some attacks don’t target the model or the inputs themselves; instead, they go after the coordination layer, which plays a major role in distributed systems~\cite{liyanage2022zsm,debenedetti2024agentdojo}. The role of the orchestration layer becomes all the more important in 6G systems, where the network relies on real-time, tight-loop interactions among user devices, edge nodes, service platforms, and multiple cooperating agents. The communication system is not just an operational backbone but also an active attack surface~\cite{he2025communicationattacks,triedman2025mascode}. Another threat in this category is the agent's identity and inter-agent trust. A malicious or compromised agent can impersonate a benign peer agent and misuse that trust relationship to inject false coordination messages. In 6G settings, these effects cascade quickly to corrupted recommendations, conflicting resource decisions, unauthorized handoffs, or poisoned shared memory. Since multi-agent systems are built on fast, continuous information exchange, even a single trust relationship violation can quickly spread across the network~\cite{triedman2025mascode,shahroz2025agentsundersiege}.
	
	Coordination disruption exploits the interconnectedness of agents even more deliberately. Attackers leverage the shared dependencies and synchronization mechanisms to turn a small local error into a system-wide disruption. For example, a compromised edge agent feeds incorrect context to a slice manager; the manager then misallocates resources, and that misallocation degrades the quality of experience for a whole cluster of services downstream. The real danger is that a local compromise can turn into a distributed failure quickly, and it’s not easy to isolate and recover from such faults easily~\cite{he2025communicationattacks,shahroz2025agentsundersiege}. Availability and resource-exhaustion attacks work by overwhelming the decision-making infrastructure rather than corrupting it. Attackers execute this by using unusually complex prompts, triggering recursive tool calls, or engineering sequences of events that keep agents perpetually busy. In 6G deployments, network latency plays a crucial role. Even a little compromise here can lead to service disruption. So, this kind of attack combines the denial-of-service threat combined with the communication-performance degradation in a single mechanism~\cite{triedman2025mascode,shahroz2025agentsundersiege}.
	
	Finally, control-loop and policy manipulation attacks target the layer where autonomous decisions are translated into actual network actions. The potential targets for this attack are - slice orchestration, digital twin synchronization, scheduling routines, and semantic service adaptation. The distinctive feature and danger of this kind of attack is that they don’t need to touch the transport layer or the actual devices. By such a quiet influence on the network governance layer, such attacks can go unnoticed for a long time~\cite{shi2025toolhijacker,hu2026maltool,xu2024llmagents6g}. On an overall basis, the study of orchestration-level threats indicates that the attack surface extends beyond the AI module to the entire distributed ecosystem of users, agents, tools, memory, and communication infrastructure. Understanding of this full picture gives motivation for the layered defense strategy discussed in the next section.
	
	\begin{figure*}[t]
		\centering
		\includegraphics[width=\linewidth]{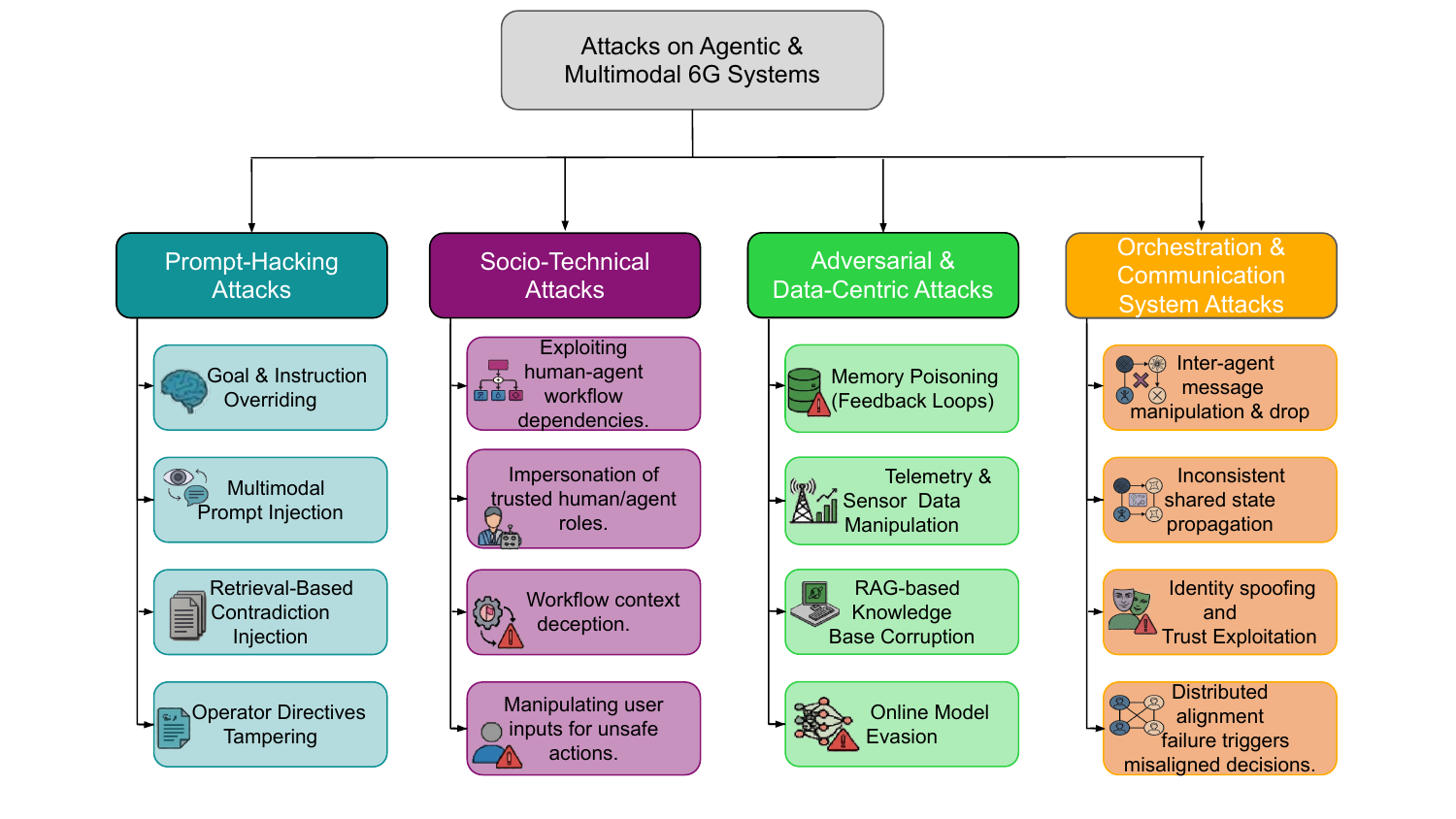}
		\caption{Taxonomy of attack categories targeting agentic and multimodal 6G systems.}
		\label{fig:6g_attack_taxonomy}
	\end{figure*}
	
	\section{Mitigation Strategies and Defense Design}
	
	The mitigation categories below similarly mix two kinds of evidence: controls already studied in the adjacent agent-security, trustworthy-AI, and communication-systems literature, and 6G-oriented design principles proposed here as plausible deployment directions rather than fully validated end-to-end solutions.
	
	\subsection{Input and Prompt Defenses}
	The first line of defense in secure agentic 6G systems lies at the point where instructions, multimodal content, and contextual information enter the autonomous pipeline~\cite{chen2024struq,debenedetti2025cameldefense}. Effective input defenses include prompt segmentation, instruction delimiting, re-tokenization, content sanitization, policy-based filtering, source-aware parsing, modality-specific validation, cross-modal consistency checking, and provenance verification for retrieved content. In practice, secure input handling should distinguish operator policies from user content, retrieved context from trusted internal state, and control instructions from descriptive data so that malicious semantic content is less likely to be treated as an executable objective.
	
	\subsection{Memory and Knowledge Defenses}
	After the inputs have been sanitized, it is important to make sure that the agent retains its learning over time. Persistent memory is a fundamental characteristic of an agent, making it essential to safeguard the memory and knowledge pipelines~\cite{park2023generativeagents,he2025llmagentsecurity}. The controls required for this are authenticated memory updates, mechanisms to track the source of information, integrity verification for external knowledge sources, and anomaly detection of how the agent's state evolves. Besides these, other safety measures include a rollback mechanism for corrupted entries, an explicit authentication mechanism for who is allowed to write to memory, what kinds of information are retained, policies for invalidating outdated knowledge, and how the system handles contradictions between fresh and stored evidence. In 6G contexts, proper implementation of these steps is important to keep reliable communication and service continuity over long-running tasks~\cite{xue2024badrag,debenedetti2025cameldefense}.

	\subsection{Tool and Action Defenses}
	Securing tool use is equally important as safe memory. Agentic systems derive their power from the ability to call external tools to enhance their functionality~\cite{schick2023toolformer,yao2023react}. So, it is important to ensure that these tools can’t be manipulated. This can be ensured by least-privilege access, capability scoping, policy-constrained execution, runtime verification, and detailed action logging. There should be a human approval step for high-risk operations, and tool outputs should be validated before being used as trusted context. In the context of 6G communication systems, these principles can be implemented as network-function isolation, API mediation, rate limiting, and the use of digital twins or shadow environments that let the agent simulate an action before committing to it. This ensures that autonomous reasoning doesn’t assume independent authority to make decisions, which can be exploited by adversaries~\cite{ruan2023toolemu,hu2026maltool}.

	\subsection{Inter-Agent Trust and Coordination Defenses}
	Multi-agent collaboration introduces a layer of risk, and controls are needed to ensure identity and coordination across agents ~\cite{rose2020zerotrust,wu2024autogen}. Agents require strong identity management, signed and authenticated messages~\cite{verma2020cbcas}, role-based communication policies, and trust-scoring mechanisms that evolve based on observed behavior. To ensure a trust relationship based on peer agent inputs, message provenance tracking, consistency checks across shared state, and well-defined handoff semantics must be implemented. For sensitive operations, it’s imperative to require cross verification from multiple agents before proceeding. This, however, has to be carefully balanced against task urgency, since higher latency may not be acceptable for time-sensitive 6G operations.~\cite{he2025communicationattacks}
	
	\subsection{Communication- and 6G-Layer Defenses}
	All the AI protections mentioned above must also include defense of the communication substrate~\cite{wang2020securityprivacy6g,masaracchia2022digitaltwin6g}. autonomous decisions significantly affect the communication layer, including slice management, scheduling, handoff behavior, semantic service adaptation, and digital twin synchronization in envisioned deployments; therefore, robust communication-layer security is an integral part of the overall defense strategy. Various defense measures that can be taken at this layer include keeping network slices isolated from each other, securing communication channels, and verifying policies across all control loops. Some other measures are trusted execution at the edge, integrity checks on telemetry and sensor data, and validating that the digital state actually matches what's happening physically. Digital twins provide added safety by allowing you to execute any action in a simulated environment before it’s pushed to live. This helps to identify and remediate failures before go-live~\cite{paul2025quantumdrltwin,li2025delayaware}.
	
	At the network level, misuse of autonomous authority or action of a silent adversary could be tracked by continuous monitoring, which includes watching for unusual orchestration behavior, unexpected oscillations in control loops, or resource usage spikes~\cite{alladi2021securing}. All these measures are especially important for 6G since an adversary doesn’t necessarily need to break a low-level protocol to cause damage. Manipulating the autonomous control layer may be sufficient to degrade latency, reliability, energy efficiency, or service continuity. Therefore, AI-layer and network-layer defenses should be co-designed rather than deployed in isolation.
	
	\subsection{Privacy-Preserving Intelligence Mechanisms}
	Alongside integrity and availability, privacy remains a cross-cutting concern for multimodal 6G intelligence~\cite{dwork2006calibrating,mcmahan2017fedavg,sai2024federated}. Relevant mechanisms include data minimization, selective memory retention, access-bounded retrieval, federated learning, secure aggregation, differential privacy, confidential inference, and encryption of stored or transmitted context. In distributed edge intelligence, these techniques must be combined with explicit memory policies governing what may be stored, for how long, for which tasks, and under what user or operator consent conditions~\cite{shokri2017membership,fredrikson2015modelinversion}.
	
	\subsection{Defense Tradeoffs and Design Principles}
	These mechanisms are effective only when they are balanced against broader system tradeoffs. Because secure agentic 6G systems operate under multiple constraints, defense design should be guided by explicit tradeoffs among security, latency, privacy, contextual utility, centralization, distributed autonomy, robustness, and computational cost. Several principles follow: defenses should be cross-layer; systems should enforce least privilege and bounded autonomy; retrieved content, memory updates, and inter-agent messages should carry provenance and trust metadata; safety-critical decisions should support verification before execution, ideally through simulation or cross-checking; and memory/privacy policies should be explicit and lifecycle-aware. Together, these principles provide a foundation for trustworthy multimodal multi-agent 6G systems while motivating the open challenges discussed next.
	
	\begin{table*}[t]
		\centering
		\caption{Attack vectors, targeted components, and defense directions in secure agentic 6G systems.}
		\label{tab:defense_summary}
		\renewcommand{\arraystretch}{1.15}
		\setlength{\tabcolsep}{4pt}
		
		\begin{tabular}{|
				>{\raggedright\arraybackslash}p{2.7cm}|
				>{\raggedright\arraybackslash}p{3.5cm}|
				>{\raggedright\arraybackslash}p{4.0cm}|
				>{\raggedright\arraybackslash}p{3.3cm}|
				>{\raggedright\arraybackslash}p{2.9cm}|
			}
			\hline
			\textbf{Attack} & \textbf{Targeted Component} & \textbf{Defense Strategy} & \textbf{6G Deployment Context} & \textbf{Residual Challenge} \\
			\hline
			Prompt injection & Prompt interface, reasoning context, retrieved content & Input segmentation, provenance checks, instruction delimiting, policy-aware parsing & Operator interfaces, intent agents, orchestration workflows & Semantic ambiguity and long-context evasion \\
			\hline
			Jailbreaking & Agent cognition, safety guardrails, tool invocation logic & Alignment hardening, policy verification, bounded autonomy, runtime approval gates & Zero-touch management, service automation, edge assistants & Multimodal and cross-agent jailbreak transfer \\
			\hline
			Tool chaining abuse & Tool outputs, API call sequences, delegated actions & Typed interfaces, least privilege, simulation-before-execution, action logging & Network orchestration, digital twins, slice control & Complex chains spanning multiple tools and agents \\
			\hline
			Memory tampering & Short-term context, long-term memory, vector stores & Authenticated updates, rollback, retrieval validation, retention policies & Personalization, incident history, digital twin state & Balancing adaptability with strict memory governance \\
			\hline
			Data poisoning & Telemetry repositories, knowledge bases, external corpora & Source validation, anomaly detection, data provenance, agreement-based retrieval & Edge analytics, fault diagnosis, semantic knowledge bases & Subtle poisoning in dynamic and distributed data streams \\
			\hline
			Model extraction & Model behavior, policy-selection logic, agent responses & Query monitoring, rate limiting, watermarking, response shaping & Exposed edge services, public APIs, agent-facing interfaces & Utility loss and collaborative extraction attacks \\
			\hline
			Identity/trust spoofing & Agent identities, message channels, delegated tasks & Signed messaging, role-bounded communication, trust scoring, state consistency checks & Multi-agent orchestration across device-edge-cloud domains & Latency overhead and dynamic trust re-evaluation \\
			\hline
			Resource exhaustion & Compute budget, control loops, orchestration APIs & Rate limits, admission control, adaptive verification, priority-aware scheduling & Edge inference, real-time orchestration, network management & Defending without harming latency-sensitive workloads \\
			\hline
		\end{tabular}
	\end{table*}
	
	\section{Open Challenges and Future Research Directions}
	
	\subsection{Secure Multimodal Fusion Under Adversarial Conditions}
	A major open challenge in secure agentic 6G systems is robustly fusing heterogeneous modalities under adversarial conditions~\cite{xie2021deepsc,shayegani2024jailbreakpieces}. Future communication systems will increasingly rely on combinations of text, audio, video, sensing, telemetry, and contextual metadata to infer user intent and adapt network behavior. However, multimodal fusion introduces a wider attack surface because an adversary can manipulate one modality while preserving apparent consistency in others, or can exploit disagreements between modalities to steer downstream reasoning~\cite{liu2024mmsafetybench,bailey2024imagehijacks}. Current defenses are still limited in their ability to detect subtle cross-modal inconsistencies, quantify modality trustworthiness, and reason about uncertainty across fused inputs. Future research should therefore focus on modality-aware trust estimation, cross-modal validation mechanisms, uncertainty-calibrated fusion, and robust inference pipelines that can gracefully detect when one or more modalities become unreliable. Strict latency and Resource constraints are required in 6G systems, which make secure multimodal fusion an AI and communication-systems challenge.
	
	\subsection{Trustworthy Multi-Agent Orchestration for Zero-Touch 6G}
	The system requires coordination among autonomous systems, even if the multimodal system is robust. Zero-touch and self-managed 6G networks will likely depend on multiple autonomous agents that coordinate across devices, edge nodes, service platforms, and core-network controllers~\cite{liyanage2022zsm,wu2024autogen}. Multi-agent architectures introduce unsolved problems in trust establishment, delegation control, policy alignment, and resilience to compromised peers. Existing multi-agent coordination methods do not fully address communication-system constraints, such as bounded latency, hierarchical control domains, and safety-critical orchestration; instead, they focus on integration with AI systems~\cite{jiang2024commllm,li2023camel}. A key research direction is the development of secure orchestration frameworks that provide authenticated delegation, verifiable task handoffs, role-bounded authority, conflict resolution, and fault containment across distributed agents. Such frameworks should also support mixed-trust environments, where some agents are more trusted than others and where control authority may shift dynamically across network domains. Future work should focus more on combining agent flexibility with high assurance and traceable communication.
	
	\subsection{Privacy-Preserving Persistent Memory}
	Besides coordination, privacy introduces a separate challenge. Persistent memory is central to agentic intelligence, as it enables continuity, personalization, and extended-horizon optimization~\cite{park2023generativeagents,dwork2006calibrating}. However, in multimodal 6G environments, retaining user preferences, interaction histories, sensing traces, and contextual metadata for extended periods can cumulatively lead to privacy leakage, even when no single interaction appears highly sensitive. Future research should emphasize privacy-enhanced memory architectures with task-restricted retention, access-bounded retrieval, encrypted memory structures, and consent-aware policies~\cite{shinn2023reflexion,chen2024agentpoison}.
	
	\subsection{Security of Semantic Communication and Intent-Aware Networking}
	Along with focusing on the network's semantics, the privacy of the memory must also be considered. 6G brings a shift from conventional bit-centric transport to semantic and intent-aware communication~\cite{xie2021deepsc,yang2025mgsc}. In such systems, the objective becomes preserving the meaning and relevance of information for downstream decision-making rather than just delivering it accurately. This creates new attack opportunities because an adversary may manipulate the semantics, context, or inferred intent without violating underlying transmission correctness~\cite{yang2025mgsc,zhang2024irssemantic}. Secure semantic communication, therefore, requires new threat models, metrics, and defenses that account for meaning-level integrity. Future research should examine how semantic distortion propagates through autonomous reasoning loops, how intent inference can be manipulated, and how trustworthiness can be quantified when communication quality is judged at the semantic rather than packet level. This area is especially important for multimodal and agentic systems, where meaning is derived from fused context and then translated into system actions.
	
	\subsection{Secure Digital Twins and Cross-Layer State Consistency}
	Those semantic guarantees become even more important when digital twins are used to validate actions. Digital twins are expected to play a major role in 6G by supporting simulation, optimization, diagnosis, and predictive orchestration~\cite{masaracchia2022digitaltwin6g}. In secure agentic systems, they may also act as safety buffers to validate candidate actions before deployment, provided their state remains synchronized with the physical network and is resilient to manipulation. Therefore, open problems include secure physical-digital synchronization, integrity verification for twin updates, robust detection of state divergence, and integration of digital twins into agentic loops as trustworthy verification layers without excessive overhead~\cite{lu2021flblockchaindt,paul2025quantumdrltwin}.
	
	\subsection{Benchmarking, Evaluation, and Reproducibility}
	To enable meaningful comparisons across studies, we need stronger benchmarks and clearer evaluation protocols. The gap here is the lack of a robust framework for working with multimodal agentic AI in the context of 6G networks~\cite{debenedetti2024agentdojo}. Most existing benchmarks address general LLMs with single modality and/or legacy network security datasets. There is a need to reflect how multimodal context, persistent memory, tool invocation, inter-agent coordination, and communication-layer dynamics interact with one another~\cite{liu2024agentbench,xie2024osworld}. Well-designed future benchmarks should consider the 6G communication setups, should incorporate multimodal data pipelines, handle edge-cloud execution environments, orchestration workflows, and traces of real security and privacy attacks. Currently, evaluation matrices work with attack success rate and detection accuracy. But these should be augmented to capture latency, QoE, service continuity, policy robustness, privacy leakage, recovery overhead, and reproducibility. We propose that any 6G-agentic security benchmark report these metrics together with the targeted threat class $\tau$ and reproducibility metadata (dataset, code, hardware, and seed), so that future studies can be compared like-for-like against the same threat instances. The community should support open protocols, shared threat-model assumptions, and consistent experimental settings to ensure better confidence in evaluation results.  
	
	\subsection{Governance, Standardization, and Human Oversight}
	The aim is to ensure governance and human oversight rather than just developing robust benchmarks~\cite{rose2020zerotrust,jiang2026agenticcommunications}. Proper governance structures should ensure accountability, operational transparency, and safe behavioral boundaries. And these become all the more important in 6G settings, where autonomous decisions can ripple across critical services, user privacy, and large-scale network infrastructure. This leads to many open questions: How should auditability be defined for multi-agent coordination? What is the right way to standardize trust and provenance metadata across heterogeneous vendors and network domains? And how can human oversight be structured to balance it with the efficiency gains automation delivers? Future standards should also tackle memory governance, autonomous tool authority, semantic-level integrity, and cross-domain agent interoperability. There is also a need for well-defined explainability tools that help explain why a decision was made, not just what the decision was. The decision reasons include which modalities were weighted, which memory entries were consulted, and which peer-agent interactions proved most influential.
	
	\subsection{Sustainable and Resource-Aware Secure Intelligence}
	The defenses must remain practical under 6G resource and efficiency constraints~\cite{long2025llm6gintelligence,yang2025mgsc}. Reconfigurable hardware has already been explored as a practical substrate for accelerating and adapting next-generation 5G communication functions~\cite{chamola2020fpga5g}. For secure agentic 6G systems, similar hardware-software co-design may be needed to execute lightweight inference, policy verification, logging, and trust-management functions within strict latency, energy, and resource budgets. Stronger defense techniques often require additional computations, monitoring, and storage, which can increase the latency, energy consumption, and deployment cost. Since 6G systems are intended to support large-scale distributed intelligence under efficiency constraints, effective security solutions that entail high operational overheads may be less practical to scale. This creates a need for a resource-aware security architecture that adapts the degree of protection to task criticality, context sensitivity, and available compute budget. Future research directions include adaptive verification pipelines, event-driven monitoring, selective use of strong privacy mechanisms, and security-aware scheduling for edge intelligence workloads. Sustainable, secure intelligence becomes increasingly important for 6G, as the system must jointly optimize reliability, privacy preservation, system resilience, and energy efficiency.
	
	\section{Conclusion}
	The various challenges discussed above reinforce the central message of this survey. This paper surveyed the emerging landscape of secure agentic AI and multimodal intelligence in 6G communication systems, with a special focus on cross-layer security and privacy for multimodal, memory-enabled, multi-agent orchestration. It presented a 6G-oriented architectural view of single-agent and multi-agent systems, a structured threat model spanning multimodal input, reasoning, memory, tools, and coordination, a unified attack taxonomy, and a corresponding set of cross-layer mitigation strategies. Overall, the survey distinguishes established attack and defense evidence from forward-looking inferences about 6G threat modeling and architecture. Security and privacy in agentic 6G systems must be treated as properties of the autonomous communication stack rather than as a means to solve the problem at the model level.
	
	As 6G systems evolve towards an architecture focusing on semantic, multimodal, and self-managed operations, progress will depend on advances in secure multimodal fusion, verifiable multi-agent orchestration, privacy-aware persistent memory, strong semantic communication, reliable digital twins, and rigorous evaluation standards. By integrating AI security, multimodal intelligence, and communication-system design along with a threat-modeling framework, this paper aims to support the development of reliable, privacy-preserving, and resilient 6G ecosystems.
	
	\bibliographystyle{IEEEtran}
	\bibliography{secure_agentic_ai_multimodal_6g_refs}
	
\end{document}